\documentclass[twoside,11pt]{geophysicsf}

\usepackage{natbib}
\usepackage{amsmath}
\usepackage{graphicx}
\usepackage{color}
\usepackage{amssymb}
\usepackage{amsbsy}
\usepackage{lineno}
\usepackage{ulem}
\usepackage{setspace}
\usepackage{hyperref}
\usepackage{jmlr2e}

\setfigdir{Fig}

\def\beq{\begin{equation}}
\def\eeq{\end{equation}}
\def\beqa{\begin{eqnarray}}
\def\eeqa{\end{eqnarray}}

\begin{document}

\ShortHeadings{Direct domain adaptation}{Alkhalifah and Ovcharenko}

%\jmlrheading{1}{2021}{1-48}{4/00}{10/00}{00-000}{Alkhalifah and Ovcharenko}

\author{\name Tariq Alkhalifah \email tariq.alkhalifah@kaust.edu.sa \\
\name Oleg Ovcharenko \email oleg.ovcharenko@kaust.edu.sa \\
\addr Physical Sciences and Engineering \\
King Abdullah University of Science and Technology \\
Thuwal 23955-6900 \\
Saudi Arabia}

\title{Direct domain adaptation through reciprocal linear transformations}

\maketitle

\begin{abstract}
We propose a direct domain adaptation (DDA) approach to enrich the training of supervised neural networks on synthetic data by features from real-world data. The process involves a series of linear operations on the input features to the NN model, whether they are from the source or target domains, as follows: 1) A cross-correlation of the input data (i.e. images) with a randomly picked sample pixel (or pixels) of all images from that domain or the mean of all randomly picked sample pixel (or pixels) of all images. 2) The convolution of the resulting data with the mean of the autocorrelated input images from the other domain. In the training stage, as expected, the input images are from the source domain, and the mean of auto-correlated images are evaluated from the target domain. In the inference/application stage, the input images are from the target domain, and the mean of auto-correlated images are evaluated from the source domain. The proposed method only manipulates the data from the source and target domains and does not explicitly interfere with the training workflow and network architecture. An application that includes training a convolutional neural network on the MNIST dataset and testing the network on the MNIST-M dataset achieves a 70$\%$ accuracy on the test data. A principal component analysis (PCA), as well as t-SNE, show that the input features from the source and target domains, after the proposed direct transformations, share similar properties along with the principal components as compared to the original MNIST and MNIST-M input features.
\end{abstract}

\begin{keywords}
Domain adaptation, Covariate Shift, Supervised learning, MNIST
\end{keywords}

%\inputdir{Fig}

\section{Introduction}
\vspace{-0.15in}

Machine learning (ML) is gaining a lot of traction as a tool to help us solve outstanding problems in image processing, classification, segmentation, among many other tasks.
Most of the applications in many fields have relied on supervised training of neural network (NN) models, where the labels (answers) are available \cite[]{Osisanwo,7780459,ronneberger2015unet,doi:10.1111/1365-2478.12865}. These answers are often available for synthetically generated data as we numerically control the experiment, or they are determined using human interpretation or human crafted algorithms applied to real data. The challenge in training our NN models on synthetic data is the generalization of the trained models to real data, as that process requires careful construction of the training set and the inclusion of realistic noise and other features of the real data. In other words, the synthetic and real data are usually far from being drawn from the same distribution, which is essential for the success of a trained NN model \cite[]{Kouw2018AnIT}. Thus, many synthetically trained NN models have performed poorly on real data. On the other hand, training on real data using supervised learning provides models that are often, at best, as good as the accuracy of the labels that were determined by humans or human-crafted algorithms (weak supervision). So the data-driven feature of machine learning, in this case, will be highly weakened \cite[]{10.1093/nsr/nwx106}.

\subsection{Related Work}
The concept of trying to bridge the gap between the training (source) and application (target) data is referred to as domain adaptation \cite[]{Kouw2018AnIT,Lemberger2020APO}. In this case, the training dataset is assumed to belong to the source domain and the application/testing data are assumed to belong to the target domain, the target of our training. The classic theory of machine learning assumes that the application (target) data of a trained model come from the same general population (sampled from the same distribution) as the training (source) set \cite[]{Kouw}. So we need the probability distribution of the synthetic (source) dataset, $P_s({\bf x_s},{\bf y_s})$, where ${\bf x_s}$ are the input features, and ${\bf y_s}$ are the labels from the source domain, to be similar to the probability distribution of the real (target) dataset, $P_t({\bf x_t},{\bf y_t})$, where ${\bf x_t}$ are the input features, and ${\bf y_s}$ are the labels for the target set. In real life, target labels are often missing or very limited, and thus, we have to assume that the training task is representative of the application (inference) task. In other words, the synthetically modeled labels reflect what we would expect in real life granted that the input features are similar. 
%
% However, though the task might be well represented by the training data, like a classification of a picture of a cat as a cat, and such data might be generated through computer simulation (of a cat), the application data based on real-world images might have slightly different features (i.e. background) as compared to the training (source) data. 
%
A representative dataset for a certain task, like the classification of a picture of a cat as a cat, might be generated from the available set of labeled images.
However, the application dataset constructed  from unlabeled real-world images might have slightly different features (i.e. background) as compared to the training (source) data. 
One category of data adaptation is referred to as subspace mapping (or more generally, alignment) in which methods aim to find a transformation, $T$, that brings together the distribution of the training (source) input features and those of the target data \cite[]{Fernando2013UnsupervisedVD,JMLR206}. Specifically, $P_s \left(T({\bf x_s})\right)=P_r({\bf x_r})$. 
This can, also, be accomplished by projecting the source and target input features to the eigenvectors of the two subspaces, then finding a transformation between these projected spaces. Such projections can be achieved by neural network embedding, in which we find the weights that minimize the distance between the distribution of the source samples embedding and the target samples one. There are many ways to find the transformation or weights to make the distributions similar including the use of optimal transport \cite[]{villani2008optimal}. Even cycle Generative Adversarial networks (GANs) are used for the purpose of learning a generator to map target input features to source ones \cite[]{Gupta2019CycleGANFS,JMLR239}. However, these methods become more difficult to apply when the dimensions of the input features are large.  The method, proposed in this paper, shares the general concept of this objective implemented in an empirical fashion.
 
\subsection{The paper's objectives}
An objective of a trained neural network model is to provide us with an output for a given input. The output that a trained model will give is based on the training it experienced, and that depends mostly on the source training set and its distribution. A trained neural network model generalizes well when the target data are represented, as much as possible, in the source data set. To help accomplish that when the application (target) data are real data, we propose, here, to inject the real data features into the synthetic data training. This can be accomplished by utilizing a combination of linear operations including cross-correlation, auto-correlation, and convolution between the source and target input features. These operations will bring the distributions of the training (source) input features, and the testing (target) input features closer to each other, which will help the trained model generalize better in the inference stage. In this paper:
\begin{itemize}
\item We propose a new transformation of the input features based on linear operations that will allow us to bring their distributions closer, we refer to this explicit process as direct domain adaptation (DDA).
\item The process is explicit and it operates exclusively in the data domain, without intervention into the training workflow and architecture. The method achieves 70$\%$ accuracy when a relatively simple model is trained on the MNIST data and applied to the MNIST-M data.
\item We use principal component analysis (PCA) to show how much the principal components of the MNIST and MNIST-M input features after DDA came closer to each other. 
\end{itemize}
The paper is organized as follows. We first describe the setup of the problem and our scope and then share the proposed linear transformations. We then share the results of testing the approach on the MNIST/MNIST-M data. Finally, we discuss the effect of the transformation on the data, and share some concluding remarks.

\section{Input features reciprocal projections}
The method proposed here is applicable mainly to supervised and in some cases of unsupervised (i.e. clustering) learning. We will focus here on supervised learning and specifically on the classification task. We will first share the setup of the problem and then describe the proposed transformations.

\subsection{Setup}
For the neural network (NN) model, we consider the input feature space $\chi$, a subset of $R^d$, where $d$ is the dimension of the input. We also consider the label space $Y$, a subset of $R^D$, where $D$ is the dimension of the output, and it can be as small as one for binary classification
or the number of classes, $K$, or as big as a multidimensional image. 
We assume we do not have labels in the target domain, and thus, we can not perform transfer learning (a form of domain adaptation). So in our case, the source data are labeled, but the target real data are not. Thus, we consider sample source data having inputs ${\bf x_s}$ and labels ${\bf y_s}$, having a probability distribution $P_s({\bf y_s},{\bf x_s})$, that is different from
the target data with inputs ${\bf x_t}$ and potential labels ${\bf y_t}$, given by a probability distribution $P_t({\bf y_t},{\bf x_t})$. We assume that the distribution difference is caused by the shift
in the data, and thus, $P_s({\bf y_s}|{\bf x_s})=P_t({\bf y_t}|{\bf x_t})$. This sort of 
difference between the source and target datasets distributions is referred to as a covariate shift \cite[]{Fernando2013UnsupervisedVD}. In this case, the issue we are addressing here is the case when the input (features) distributions for the source and target data are not the same, specifically
$P_s({\bf x_s}) \neq P_t({\bf x_t})$.

There are many ways to measure such a shift, including using the Kullback-Leibler (KL) divergence metric. In domain adaptation, and similar to error bounds defined for machine learning in general, we can define an error bound on the  application of a trained network model. This error bound is given by the error in the training and a term related to the complexity of the NN model (like its size). This, however, assumes that the training and application data come from the same distribution. For the case of a covariate shift, we have a similar bound, given by \cite[]{Lemberger2020APO}:
\beq
\varepsilon_t(\mathcal{NN}) \leq \varepsilon_s(\mathcal{NN}) + d(P_s({\bf x_s}),P_t({\bf x_t})) + \lambda,
\label{eq:eq22}
\eeq
where $\varepsilon_s$ is the bound on the training error, and $d(.,.)$ is the distance between the marginal distributions of source and target datasets. Here, $\lambda$ represents the optimal joint errors of the neural network model between the source and target datasets. So the upper bound of the application error is guided by these three terms.

\subsection{Our goal}
% This paragraph is very similar to one from above
One category of data adaptation is referred to as subspace mapping (or more generally, alignment). We find a transformation, $T$, that results in the distribution of the training source input features to equal that of the testing ones \cite[]{Fernando2013UnsupervisedVD}.
Specifically, $P_s \left(T({\bf x})\right)\approx P_r({\bf x})$. Compared to previous transformations, the approach used here does not require any eigenvalues computation,
which can be expensive for large dimensional input features.

So our objective is to devise a transformation that minimizes the difference measure $d$ in equation~\ref{eq:eq22} between the distributions of the training (source) and testing/application (target) input features. Specifically, we aim to find the transformation ${\bf \overline{x}_s}=T_s({\bf x_s})$ on the source input features and ${\bf \overline{x}_t}=T_t({\bf x_t})$ on the target input features so that the probability distributions $P_s({\bf \overline{x}_s}) \approx P_t({\bf \overline{x}_t})$. Figure~\ref{fig:diagramMNIST} summaries that within the framework of the training process. So the input to the training of the neural network ($\mathcal{NN}$) model is ${\bf \overline{x}_s}$, in which the model parameters are optimized to match the labels ${\bf y_s}$ using a loss function ($\mathcal{L}$). On the other hand, the input during inference is ${\bf \overline{x}_t}$. We will discuss the transformations $T_s$ and $T_r$ in the next subsection.

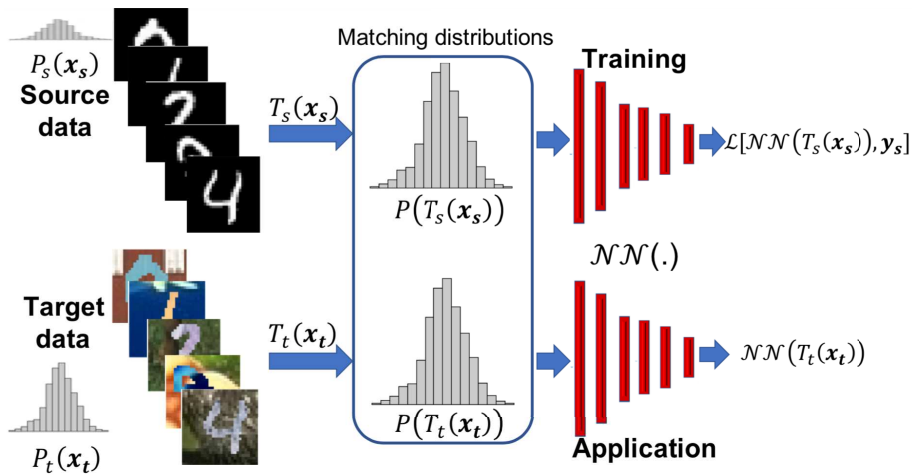
\plot{diagramMNIST}{width=0.8\textwidth}{The motivation behind the proposed data adaptation in which the training (source) dataset might have a different distribution than the application (target) dataset. Transformations $T_s$ and $T_t$ will help reduce such difference and provide new input features to the neural network function ($\mathcal{NN}$) to train the network to reduce loss $\mathcal{L}$, and then apply to real data. Here, $P$ is the probability distribution and we show schematic versions of it for the source and target data, given by samples of the MNIST and MNIST-M datasets, respectively.}

\subsection{Transforming the input features}
\vspace{-0.15in}

Without loss of generality, we focus our development on two-dimensional input features, like images. The approach is applicable to one or three-dimensional input features
by using the corresponding one or three dimensional convolution and correlation operations.

Images in general, including pictures and medical scans, can often be represented by a combination of reflectively, ${\bf r}$, source (like source of light), ${\bf s}$, and noise ${\bf n}$, as follows:
\beq
{\bf x_t^i} = {\bf r^i} \ast {\bf s^i} + {\bf n^i}.
\label{eq:eq1}
\eeq
Depending on the data, all three components $\left({\bf r},{\bf s},  {\bf n} \right)$ can vary over input target features (indexed by ${\bf i}$), and have their own distributions. For ML training, we often use 
synthetically generated images, where the labels are known, considering the data are more likely simulated. To help improve the generalization of such trained NN models on images represented by equation~\ref{eq:eq1}, we often try
to include a proper representation of these components from the target input features into the synthetic (source) input features. However, in many cases, this is hard and costly.
 and that will effect the generalization of the trained model. For one, our synthetic data are often free of noise or we add random noise to them.
So we propose here operations that will transform information between the source and target input features. 

To migrate the components described in equation~\ref{eq:eq1} from the target input data to the source, we use the following linear operations to obtain the transformed input source features:
\beq
{\bf \overline{x}_s^i } = T_s({\bf x_s^i}) = {\bf c_s^i} ({\bf x_s^i }) \ast {\bf \overline{a}_t} ({\bf x_t }),
\label{eq:eq2}
\eeq
given as  a convolution of representations from the source and the target input features. Specifically, to properly scale the input source features for training, we correlate them with randomly drawn pixel (or pixels) values, ${\bf \tilde{x}_s^j}$, from input source image indexed by $j$, as follows:
\beq
{\bf c_s^i} ({\bf x_s^i }) = {\bf x_s^i} \otimes {\bf \tilde{x}_s^j}
\label{eq:eq23}
\eeq
The random index $j$ varies per input image and per epoch to allow for proper representation of the pixels in the training in which the input images are correlated
with the operator $\otimes$ representing cross-correlation. 
%Thus, the mean $({\bf \tilde{x}_s})$ of these randomly chosen pixels over the total number
%of input features (images), $N_s$, is correlated with the input feature (image), with the operator $\otimes$ representing cross-correlation. 
This operation amounts to scaling the input features with representations of the source domain. 
The output from this operation
is then convolved with the mean of the auto-correlated target data input features, given by
\beq
{\bf \overline{a}_t} ({\bf x_t }) = \frac{1}{N_t} \sum_{\bf j}^{N_t} {\bf x_t^j} \otimes {\bf x_t^j},
\label{eq:eq24}
\eeq
where $N_t$ is the number of input features (images) in the target domain. The auto-correlation here, considering 2D input features, is a two-dimensional one and is
applied in the Fourier domain, along with the convolution. In the Fourier domain, these operations reduce to simple multiplications. The mean of the auto-correlation 
of the input features provides a measure of the distribution of the data. For example,  the auto-correlation of random noise yields a quasi delta-function at zero lag proportional to the energy of the noise. A convolution
with such a function will incorporate that energy into the synthetic data so that the signal-to-noise ratio (SNR) in the transformed source input features
would be comparable to that of the auto correlated target data.
To allow for the transformed input features from the target domain to have a similar distribution to the transformed input features from the source domain, we apply a similar transformation
to that in equation~\ref{eq:eq2} with the roles reversed.

As such, we apply the following transformation on the target input features during the inference stage:
\beq
{\bf \overline{x}_t^i } = T_s({\bf x_t^i}) = {\bf c_t^i} ({\bf x_t^i }) \ast {\bf \overline{a}_s} ({\bf x_s }),
\label{eq:eq3}
\eeq
which includes the same operations as in equation~\ref{eq:eq2} with the role of the source and target input features reversed. Thus,
\beq
{\bf c_t^i} ({\bf x_t^i }) = {\bf x_t^i} \otimes {\bf \tilde{x}_t}  = {\bf x_t^i} \otimes \frac{1}{N_t} \sum_{\bf j}^{N_t} {\bf \hat{x}_t^j},
\label{eq:eq33}
\eeq
where, now, ${\bf \hat{x}_t^j}$ represent randomly drawn
sample pixel (or pixels) from the input features from the target data to scale them with the mean, ${\bf \tilde{x}_t} $.
Also, the mean of the auto-correlation of the input features from the source data is given by
\beq
{\bf \overline{a}_s} ({\bf x_s }) = \frac{1}{N_s} \sum_{\bf j}^{N_s} {\bf x_s^j} \otimes {\bf x_s^j},
\label{eq:eq34}
\eeq
where $N_s$ represents the total number of input features (images).
The idea behind having two instances of the input features from each domain in equations~\ref{eq:eq23},~\ref{eq:eq24},~\ref{eq:eq33}, and~\ref{eq:eq34},
is to balance their contribution in the new source and target domains.  This way, we match properties of the input features from the source and target domains used for training and inference, respectively,
and yet maintain the critical-for-classification features of the original input features mainly intact. 
Note that the convolution operation that connects equal amounts of the source and target
contributions in equations~\ref{eq:eq2} and~\ref{eq:eq3} is commutative. We will see the value of this operation more in the next section. Meanwhile, Figure~\ref{fig:traintestDataMNIST} (left)
demonstrates the process of applying equation~\ref{eq:eq2}, where the input features correspond to the MNIST original labeled data. On other hand, Figure~\ref{fig:traintestDataMNIST} (right)
demonstrates the process of applying equation~\ref{eq:eq3} on the MNIST-M data, which we assume to be label-free (the labels used for accuracy assessment). Note that the resulting images after transformation of the same digit look similar for the
two processes. Since this approach is relatively direct, requiring no optimization or eigenvalue projections, we refer to it as direct domain adaptation (DDA).

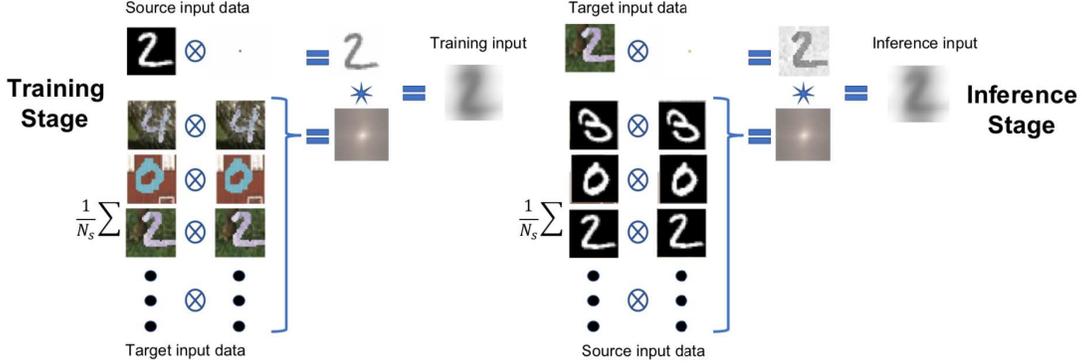
\plot{traintestDataMNIST}{width=0.95\textwidth}{The workflow chart for the proposed data adaptation applied as an example to the MNIST/MNIST-M datasets. On the left, the proposed process is used for producing the training data; on the right, the proposed process is used for producing the testing/application data. 
The circled cross symbol denotes a cross-correlation operation, and the star symbol denotes a convolution operation.}

\subsection{A Fourier domain analysis}
If we transform the target images to the Fourier domain, considering the basic laws of the Fourier representation of cross-correlation and convolution, equation~\ref{eq:eq1} can be written as
\beq
X_t^{i} = A^{i} e^{I \phi^{i}} = R^{i} S^{i} + N^{i},
\label{eq:eq4}
\eeq
where $A$ and $\phi$ are the amplitude and phase, respectively, of the complex-valued Fourier representation, with $I$ representing the imaginary unit. All capital letters indicate the Fourier representation form of the reflectivity, source, and noise functions in equation~\ref{eq:eq4}, given respectively. As a result, we can write equation~\ref{eq:eq2}
in the Fourier domain as
\beqa
X^i_s &=& \bar{X_s}^i  \tilde{X_s}^j \left( \frac{1}{N_r} \sum_j  \bar{X}^{j} X^{j} \right)= \bar{X_s}^i \tilde{X_s}^j \left( \frac{1}{N_r} \sum_j \left(A^{j}\right)^2 \right), \nonumber \\
   &=& \bar{X_s}^i \tilde{X_s}^j \left(\frac{1}{N_r} \sum_j \left(R^{j} S^{j}+ N^{j}\right) \left(\bar{R}^{j} \bar{S}^{j}+ \bar{N}^{j}\right) \right),
\label{eq:eq5}
\eeqa
where the overstrike, $\bar{.}$, symbol here stands for the complex conjugate. Note that the transformed source input features will contain elements of the target noise and
reflectivity. More importantly these elements do not affect the shape of the features as they have zero phase.

The application of the model on target data will involve an input to the model given by equation~\ref{eq:eq3}, which can be represented in the Fourier domain by
\beqa
X^i_r &=& \bar{X}^i \left(\frac{1}{N_r} \sum_j \hat{X}^j \right) \left( \frac{1}{N_s} \sum_j \bar{X_s}^{j} X_s^{j} \right) \nonumber \\
                &=& \left(\bar{R}^{i} \bar{S}^{i}+ \bar{N}^{i}\right) \left(\frac{1}{N_r} \sum_j^{N_r} \left(\hat{R}^{j} \hat{S}^{j}+ \hat{N}^{j}\right)\right) \left(\frac{1}{N_s} \sum_j \bar{X_s}^{ij} X_s^{ij} \right),
\label{eq:eq6}
\eeqa
where $\hat{R}$, $\hat{S}$, and $\hat{N}$ correspond to the aforementioned random pixel reflectivity, source, and noise, respectively, expressed in the Fourier domain. In this domain, for a single random pixel case, these quantities are constant over the Fourier domain, and the summation will render an average over the target input features. 
If the random pixel is placed in the middle as we do here, then all these quantities are real admitting no phase shifts, like in the case of auto-correlation in equation~\ref{eq:eq5}. So from equations~\ref{eq:eq5} and~\ref{eq:eq6}, we clearly notice that DAA given by both transformed input features contain equal representations of the source, reflectively, and noise from the target domain.
Since, the two data sets, after transformation,
share the same elements sampled from their corresponding distributions, then the distributions of the two data sets should be close, which allows us to satisfy the requirement
$P_s(T_s({\bf x})) \approx P_s(T_s({\bf r}))$ for the generalization of the NN model. In other words, the second term in equation~\ref{eq:eq22} will be small.

%\subsection{The distributions of the training and application data sets}

%We next evaluate how much did we manage to bring the the training (source) and testing/application (target) datasets closer to each other with the proposed transformations.
%So the
%underlying assumption in supervised learning, which is a statistical exercise in mapping inputs to outputs, where inputs and outputs correspond to particular distributions, is that the %training and application data (at least inputs) should fall within the same feature space, or in other words, come from the same distribution. In machine learning, domain adaptation is
%often handled through machine learning procedures, including cycle GAN and transfer learning. Here, considering we are dealing with waveforms in two dimensional (even 3D) space. we perform the transformation through 

%The key operation used here to tie the two data is the convolution process, which involves a multiplications of elements and since the multiplication process is cumulative,
%then the distribution of the new data is given by
%\beq
%\overline{P}_s(z) = \int P_s(x) P_t(\frac{z}{x}) \frac{1}{|x|} dx = \int P_t(x) P_s(\frac{z}{x}) \frac{1}{|x|} dx = \overline{P}_r(z)
%\eeq

\section{The MNIST data example}
\vspace{-0.15in}

The MNIST dataset has commonly been used to evaluate the performance of supervised learning algorithms tasked to classify images \cite[]{726791}. The image dataset contains 60000 training samples and 10000 testing samples of handwritten digits,
with their corresponding labels (the digits). Training accuracy has reached almost 100$\%$ \cite[]{byerly2021routing}.
As a follow-up, the MNIST-M dataset was developed as a more complicated form of MNIST. The images of the MNIST digits were combined with patches randomly
extracted from color photos and thus have 3 color channels rather than monochrome images in the MNIST dataset. The MNIST-M dataset has been used to test newly proposed domain adaptation methods \cite[]{ganin2016domainadversarial,10.5555/3045118.3045130}. 
%It include 90000 test images. In this example, we will consider the original MNIST data as our training (source) data, and we will try to adapt the NN model to the MNIST-M
It is made up of 59000 training images and 9000 testing images, which we assume to be labelless.

%The proposed transformations take place in the data domain, thus the method does not require any intervention into network architecture.

\subsection{The setup}
As mentioned above, the target MNIST-M input features are colored images described over 3 RGB channels, whereas the available training data inputs are gray-scale. We transfer features of the source dataset to the target dataset, and vice-versa, by a sequence of linear operations explained earlier. The transformed source dataset then is used for training the network while the validation happens on the transformed target dataset. Since the MNIST data are single-channel gray-scale images, we simply replicate the mono-channel values into the three channels. 

The architecture of the deep neural network for digit classification is inherited from \cite{ganin2016domainadversarial}. In this paper, the authors introduce the concept of domain adversarial neural networks for domain adaptation and design the generative-adversarial architecture (DANN) for this purpose. The generator used in their digit recognition  example originally produces two outputs - categorical for digit labels and binary to identify the domain of the input data. We modify this generator by keeping only categorical output, which assigns labels from 0 to 9 for the input images. The resulting convolutional network (Figure~\ref{fig:arch}) is a variant of the classic LeNet-5 architecture for digit recognition \citep{lecun1998gradient}. Specifically, the convolutional part of the network is built as a stack of two blocks where each block consists of a convolutional layer,  batch normalization, and max pooling, followed by a ReLU activation function. The classification head is built as a set of fully-connected layers and batch normalizations followed by ReLU. The output from the last fully-connected layer is passed through the Softmax activation function that is equivalent to assigning probabilities among categorical labels. There are also two Dropout  layers placed in the encoder and classification head. 

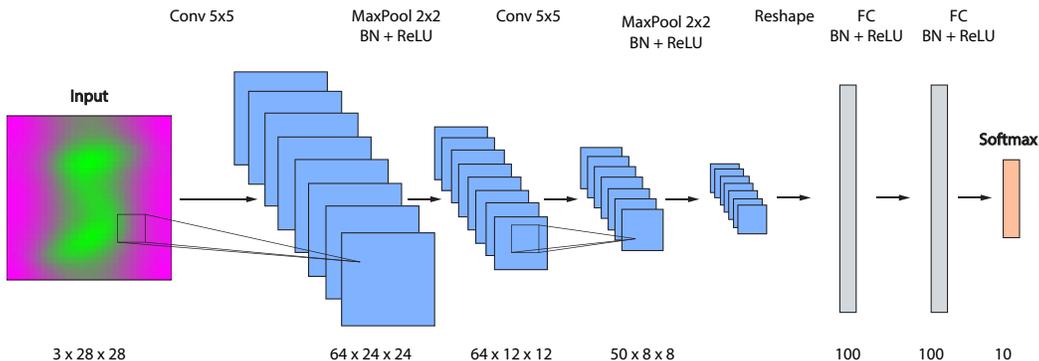
\plot{arch}{width=0.9\textwidth}{The architecture of the convolutional neural network for digit classification. The proposed method operates in the data domain and does not require intervention into the existing training workflow.}

\subsection{Adaptation process}

We apply the DDA to the MNIST and MNIST-M data using equations~\ref{eq:eq2} and~\ref{eq:eq3}, respectively.
In Figures~\ref{fig:img04F2} and~\ref{fig:img59F2}, we show the effect of DDA on an example image for each of the 10 digit classes. The left column of each Figure shows
the training (source) samples transformation, while the right column shows the inference (target) samples. For each digit, we show the color image (top row), and the corresponding individual RGB channels (bottom row), which constitute the actual input to the classification network. The images labeled ${\bf c_s}$, ${\bf c_t}$, ${\bf \overline{a}_s}$, and ${\bf \overline{a}_t}$
correspond to equations~\ref{eq:eq23},~\ref{eq:eq33},~\ref{eq:eq24}, and~\ref{eq:eq34}, respectively. Note that the final transformed input features for the source and target
data look very similar. This similarity is more profound in the 3-channel representation, which are the actual inputs. We also, however, ended up with a reverse in polarity apparent in the input features for digits 0, 3, 5, and 8.
This will be covered by the polarity reversal we will implement to augment the training (source) data, and such a reversal is demonstrated in the Figure, as well.

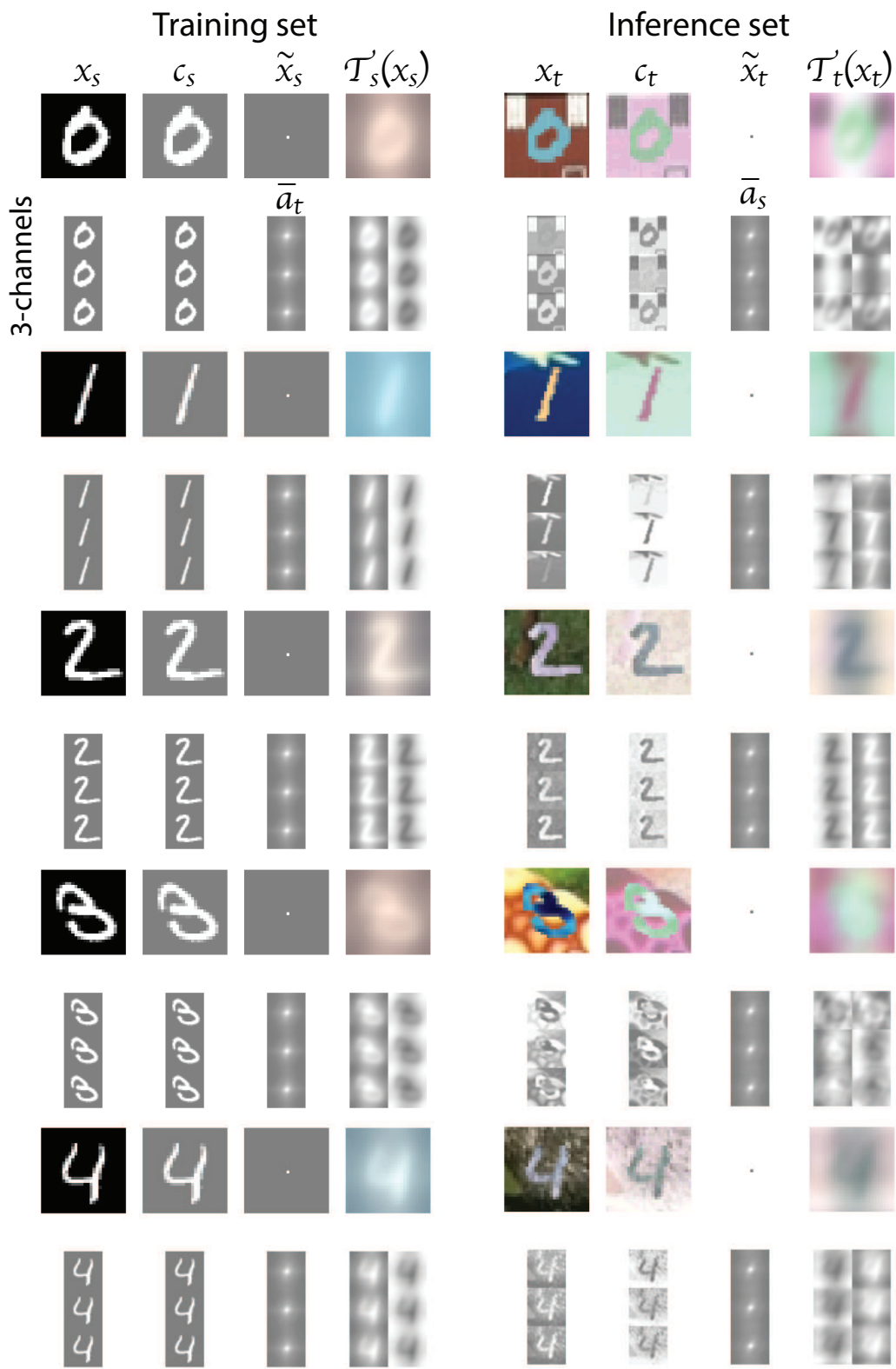
\plot{img04F2}{width=0.9\textwidth}{The DDA transformation applied to gray-scale and colored digits from 5 to 9. The left column shows the transformation of source (MNIST) data, and the right column shows the transformation of the target (MNIST-M) data. For each input, we show the composite and the 3-channel transformation. The last column of the second row shows the 3-channel output and the same with reversed polarity, which we use as the augmentation of the training data.}

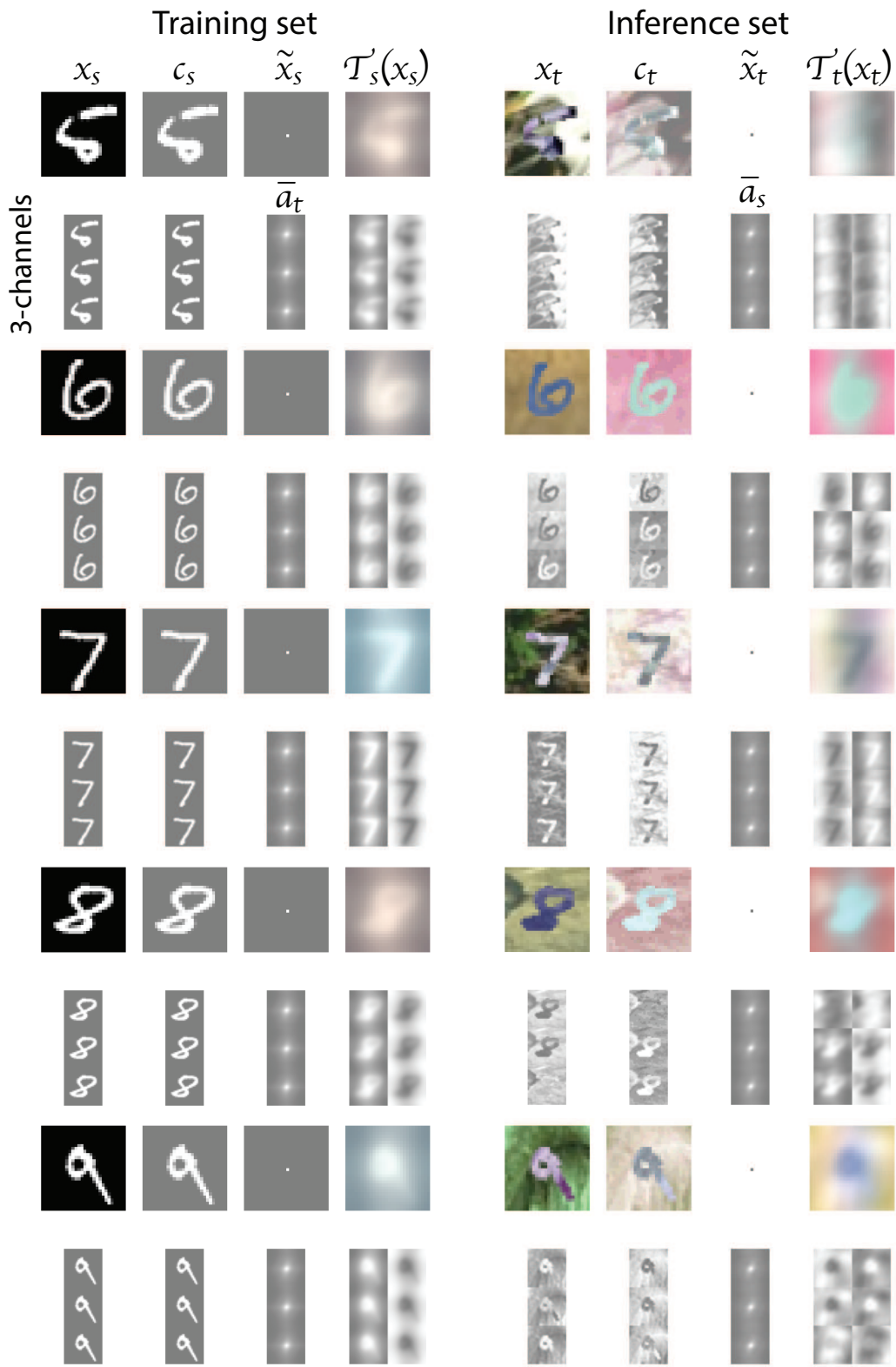
\plot{img59F2}{width=0.9\textwidth}{Same as Figure~\ref{fig:img04F2} but for digits in range from 5 to 9.}

\subsection{Data augmentation and training}
We apply DDA to the  input training data on the fly. Specifically, given a sample of  input data from the source domain (grayscale image),  we first cross-correlate it with a random pixel drawn from another arbitrary sample from the same dataset. Then, we convolve the result with the mean of the auto-correlation of the entire target dataset and re-normalize the outcome to fit the [-1, 1] range. The re-normalization is trivial and implies subtracting from the input features the minimum value among all channels, then dividing by the maximum of the previous result, which will bound the values between zero and one, and finally multiplying the output by 2 and subtracting 1. 

At the inference stage, given a sample  from the target dataset, we cross-correlate it with the average of pixel values from the same dataset (specifically, we randomly draw a pixel from each image in the dataset and compute their average) and convolve the result with the average of auto-correlation of the source dataset. For inference on the sample from the source dataset, we would reverse domains and first cross-correlate the sample with the mean of pixel values from the source dataset, followed by the convolution with the mean auto-correlation of the target dataset.\\

We train the network for 100 epochs using a batch size of 128 and an Adam optimizer \citep{kingma2014adam} with the learning rate set to $1e-3$  . To emphasize the capability of our proposed DDA approach, we augment the data using trivial methods. These include random polarity reversal of color channels and channel-shuffling. The first one compensates for the polarity mismatch between transformed images in the two datasets so we train the network to deal with it. The second augmentation effectively  expands the dataset by changing the order of channels in the image.

The inference accuracy on the transformed test partition of the  source dataset reaches 99$\%$, similar to the scenario when the network was trained on the original data from the MNIST dataset. This suggests that the features of the source dataset are not distorted to the point in which the score degrades. It still maintains the resolution of the original MNIST. We also show the performance of the DANN approach mentioned earlier for comparison. This method reaches 90$\%$ accuracy on the colored dataset by minimizing the domain gap during the adversarial training.  The proposed DDA method operates in the data domain and reaches an accuracy of more than 70$\%$ for inference on the target MNIST-M dataset.  Meaning that the DDA might be integrated into the  existing training routine without changing the network architecture. Finally, we note that the classification accuracy score reached by DDA is higher than what we would obtain without the proposed transformation of 30$\%$ (Figure~\ref{fig:accuracy_curve}). 

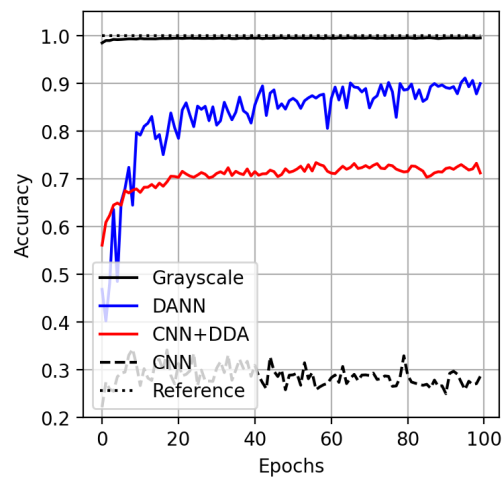
\sideplot{accuracy_curve}{width=0.75\textwidth}{The accuracy curves for application of trained networks on test partition of MNIST (solid black)  and MNIST-M (other) datasets. The inference score on MNIST-M by CNN (dashed), DANN (blue), and the proposed method (CNN + DDA, red). The perfect score is shown as the dotted line.}

\subsection{Analysis}
To further analyze the proposed direct domain adaptation approach, DDA, we display the principal component analysis (PCA) for the source and target input features before and after the transformations~\ref{eq:eq2}
and~\ref{eq:eq3} (Figure~\ref{fig:pca_circles_crosses}). Prior to DDA, the source (crosses) and the target (circles) input features occupy two different areas and they are separated mainly by the first principal component (the horizontal axis), which accounts for mainly the background of the images. After DDA, the source and target input features reasonably overlap and they follow certain slopes that are controlled by features in the images. A closer look is provided by Figures~\ref{fig:pca_images_before} and~\ref{fig:pca_images_after} in which the actual images before and after DDA are shown, respectively, as a function of the principal components. Along these slopes, which are linear combinations of the two principal components (eigenvectors corresponding to the two largest eigenvalues), we notice the images from the source and target data having, more or less, similar backgrounds. The source images (dashed box in Figure~\ref{fig:pca_images_after}) gravitate toward the edges of the slopes, while target images dominate the middle. In other words, the source images after DDA have a larger magnitude of the principal components, or in other words, more pronounced features along the principal components, which could be helpful for the training.

We further analyze the effect of the transformation using t-SNE, which exposes nonlinear features in the images. We first compare the t-SNE for the MNIST (source) and MNIST-M (target) images before and after DDA (Figure~\ref{fig:tsne_circles_crosses}). The separation between the two domains is more obvious in this plot.
Thus, an NN model trained on the source data will have a hard time producing accurate classification on the target (MNIST-M) data due to the obvious covariate shift. This was reflected in the low $30\%$ accuracy we got in applying the MNIST data trained model on the MNIST-M dataset. After applying DDA, we see clusters of mixed source and target data. A closeup look in Figures~\ref{fig:tsne_images_before} and~\ref{fig:tsne_images_after} reveal that these clusters are defined by the new general background of these images (which is a mix of the original backgrounds). Luckily, these clusters include both data domains. This implies that the transformation
formed new shared backgrounds between both data (source and target), which explains the improved accuracy in the application of the trained model on the target data.

\plot{pca_circles_crosses}{width=0.85\textwidth}{The PCA plots of the MNIST (circles) and MNIST-M (crosses) images with the corresponding digits they represent
given in a unique color before (left) and after (right) the proposed transformations.}

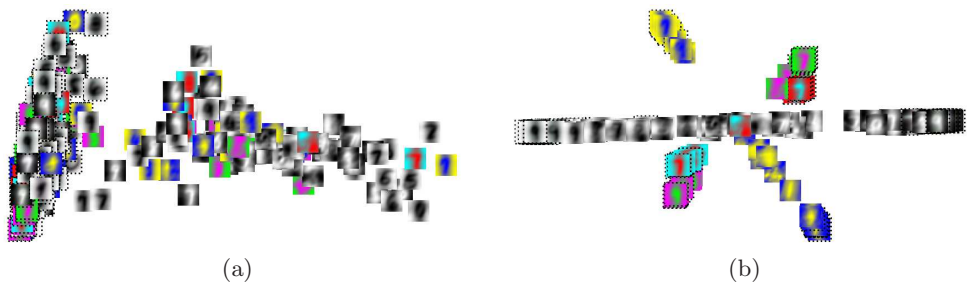
\multiplot{2}{pca_images_before,pca_images_after}{width=0.4\textwidth}{The PCA result of Figure~\ref{fig:pca_circles_crosses} with the input images shown
 a) before and b) after the proposed transformation.}

\plot{tsne_circles_crosses}{width=0.85\textwidth}{The t-SNE plots of the MNIST (circles) and MNIST-M (crosses) images with the corresponding digits they represent
given in a unique color before (left) and after (right) the proposed transformations.}

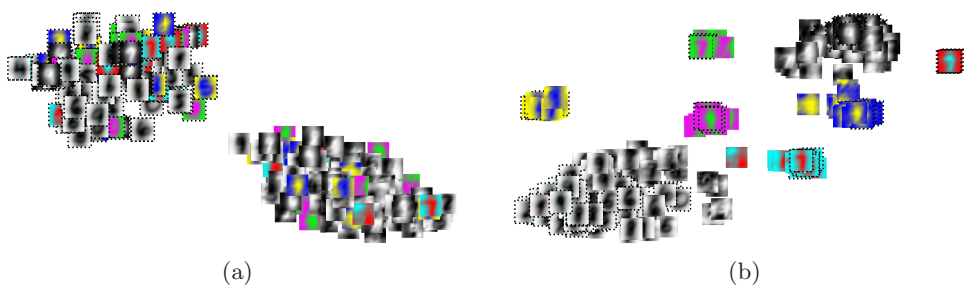
\multiplot{2}{tsne_images_before,tsne_images_after}{width=0.4\textwidth}{The t-SNE result of Figure~\ref{fig:tsne_circles_crosses} with the input images shown
 a) before and b) after the proposed transformation.}

\section{Discussions}
\vspace{-0.15in}
The transforms proposed to the input features from the source and target data are linear, explicit, and efficient to apply. The mean of the  auto-correlation (${\bf \overline{a}_s}$, 
${\bf \overline{a}_t}$) and the mean of the randomly drawn pixel/pixels (${\bf \hat{x_s}}$, ${\bf \hat{x_t}}$) can all be computed once and then applied to the input features like a linear layer. The Fourier domain provides a venue for an efficient application of these terms. Unlike conventional domain adaptation methods, the process here requires no training. Since these operations are linear, there is generally no loss in resolution in the input features, as these processes translate to multiplications in the Fourier domain. So the transformed input features carry more or less the same information as the original features. This assertion is supported by the training accuracy we achieved on the transformed MNIST data, which is at the same level as the original MNIST data using the same network.

Since DDA produces input features with equal amounts of source and target information, the distributions of the new transformed input features are expected to be closer. DDA also does not alter the general shape of the input feature, and thus, is expected to be applicable to not only classification problems, but also, segmentation, as well as other supervised learning tasks in general.

The mean of the auto-correlation of the input features is meant to extract the energy distribution information from one domain and insert it into the other domain through the convolution process. If the target domain includes a single example, the auto-correlation will contain more information on the energy distribution of that single example, and that information will be embedded in the training set. Recall that the autocorrelation process admits zero phase, and thus, does not alter the general features of the input image. For example, if the target set only includes  one sample given by the digit 7, the autocorrelation will capture the background of this single image, and the shape of this digit will be equally distributed on all the training set, and the differences given by the images of the digits will be maintained.

The PCA and t-SNE of the input features before and after DDA revealed to us that the approach managed to bring the input features (source and target) closer to each other, at least along the principal components. For the MNIST and MNIST-M data, the transformations produced a shared background for the input features that has elements from the original backgrounds of the source and target input features.

\section{Conclusions}
\vspace{-0.15in}

We proposed a direct and explicit technique to precondition the input features for a supervised neural network optimization so that the trained model works better on
available label-free target data. This concept of direct domain adaptation (DDA) is based on incorporating as much information from the target input features into the training without harming the source input features crucial for the prediction. Considering the two domains (source and target), we specifically cross-correlate an input section from one domain with the mean of randomly picked samples from the same domain followed by a convolution with the mean of the auto-correlated sections from the other domain. For training the NN model, the input image is from the source domain, and for the application, the input image is from the target domain. The DDA operates in the data domain, and thus, does not require changes in the existing network architecture and logic of the training workflow.  Thus, it might be incorporated into existing  solutions as a component of the data pre-processing routine.  A test of this approach on the MNIST data as a source and the MNIST-M data as target admitted 70$\%$ accuracy on the target data.

\section{Acknowledgments}
\vspace{-0.15in}
%%%%%%%%%%%%%%%%%%%%%%%%%
We thank KAUST for its support and the seismic wave analysis group (SWAG) for constructive discussions. The results in this paper can be accessed and reproduced through the
following GitHub link \url{https://github.com/swag-kaust/dda_pytorch}.

\bibliography{ms}

\end{document}